\documentclass[twoside,11pt]{article}

\usepackage{jmlr2e}
\usepackage{jmlr2e}
\usepackage{amsmath}
\usepackage{booktabs}
\usepackage{multirow}
\usepackage{pgfplots}
\pgfplotsset{compat=1.18}

\makeatletter
\@preprinttrue
\providecommand{\@editor}{}
\makeatother

\definecolor{cConst}{RGB}{110,110,110}
\definecolor{cStep}{RGB}{230,140,20}
\definecolor{cExp}{RGB}{30,170,170}
\definecolor{cCos}{RGB}{40,150,60}
\definecolor{cWarm}{RGB}{50,90,200}
\definecolor{cBrach}{RGB}{205,40,45}

\newcommand{\etamax}{\eta_{\max}}
\newcommand{\etamin}{\eta_{\min}}
\newcommand{\method}{BrachistoneLR}

\usepackage{lastpage}

\ShortHeadings{A Brachistochrone-Inspired Learning-Rate Schedule}{Rahman Roni, Chowdhury and Dash Nimi}
\firstpageno{1}

\begin{document}

\title{\method: A Brachistochrone-Inspired Learning-Rate Schedule and a Controlled Benchmark of Scheduling Policies}

\author{\name Md. Sadekur Rahman Roni \email sadekur\_cse@lus.ac.bd \\
       \addr Department of Computer Science and Engineering\\
       Leading University\\
       Sylhet, Bangladesh \\
       \name Md. Jalal uddin Chowdhury \email jalal\_cse@lus.ac.bd \\
       \addr DeepNet Research and Development Lab\\
       Sylhet 3100, Bangladesh \\
       \name Moutusi Dash Nimi 
       \email mou07nimi@gmail.com \\
       \addr Department of Electrical and Electronic Engineering\\
       Leading University\\
       Sylhet, Bangladesh}

\editor{Editor Name}

\maketitle

\begin{abstract}%
The learning-rate schedule is a consequential choice in training deep networks, yet the policies in
common use are heuristic and published comparisons are hard to read, because architecture, dataset and
budget tend to vary alongside the schedule. We study \method{}, a schedule built by mapping the vertical
coordinate of the \emph{brachistochrone}, the curve of fastest descent under gravity, onto the range
between a peak and a floor rate. Expanding the definition shows it to be cosine annealing with the
half-period set to $E-1$ instead of $E$, the configuration a standard implementation gives when its
period argument is one less than the number of epochs. The rate therefore reaches its floor at the last
epoch trained rather than one epoch later, and we show this difference decays as $E^{-2}$, making it a
short-horizon effect. We then benchmark six schedules over $72$ runs on three image-classification
datasets (MNIST, Fashion-MNIST, CIFAR-10) and four architecture families (fully connected,
convolutional, recurrent, residual), fixing the optimizer, data pipeline and evaluation protocol so that
only the schedule varies. Schedules that fall smoothly from peak to floor beat the constant rate and
calendar-based decay by margins that grow with task difficulty, reaching $2.5$ points of dataset mean on
CIFAR-10. Within that leading group, \method{}, cosine annealing and warmup-cosine lie within $0.06$
accuracy points and $0.17$ of a mean rank, which one seed per configuration cannot separate. \method{}
is best on both residual networks and has the highest CIFAR-10 mean, and it sets no milestones, decay
factor, warmup length or restart period. We conclude that the shape of a schedule matters more than its
parameterization, that the choice of whether to use a smooth schedule matters more than the choice among
them, and that the terminal-rate distinction is worth attention only over short horizons.
\end{abstract}

\begin{keywords}
  learning rate scheduling, deep learning optimization, cosine annealing, image classification,
  empirical benchmark
\end{keywords}

\section{Introduction}
\label{sec:intro}

The learning rate is usually the first hyperparameter a practitioner tunes and often the one that
matters most \citep{goodfellow2016deep}. Its initial value is only part of the story: the rule by which
it is varied during training affects both how quickly optimization escapes a poor initialization and how
well the resulting solution generalizes. A large rate early helps the iterates leave regions of high
loss, while a small rate late is what makes the final approach to a minimum stable, and any useful
schedule has to reconcile these two requirements \citep{loshchilov2017sgdr,smith2017cyclical}.

The policies in common use reconcile them in different ways. A constant rate is the simplest choice but
leaves late-training refinement entirely to the optimizer. Step and exponential decay reduce the rate on
a fixed calendar and oblige the user to choose milestones or a decay factor. Cosine annealing follows a
half-cosine from a peak to a floor and has become a strong default \citep{loshchilov2017sgdr}, and
warmup-cosine prepends a short linear ramp that stabilizes the first updates of large or deep models
\citep{goyal2017accurate}. Because these policies differ both in shape and in the number of quantities
they expose, and because published comparisons are drawn from experiments that vary architecture,
dataset and budget at the same time, it is difficult to attribute any reported difference to the
schedule itself.

We do three things in this paper. We introduce \method{}, a schedule whose shape is taken from the
brachistochrone, the classical curve of fastest descent. We then characterize its relationship to cosine
annealing exactly, and the characterization is deflationary: \method{} is not a new family of schedules
but a particular choice of period within an existing one, obtainable in any standard library by setting
the cosine period to $E-1$. We state this at the outset because it determines what the rest of the paper
can and cannot claim. Finally we report a benchmark of six schedules over four architectures and three
datasets ($72$ runs) under a single fixed protocol, so that the schedule is the only thing that changes
from run to run. The value of the paper lies in the second and third of these rather than the first: a
closed-form account of what the period choice does, and controlled evidence on how much the choice of
schedule is worth in the first place.

\section{Related Work}
\label{sec:related} 

Cosine annealing was introduced as a component of SGDR \citep{loshchilov2017sgdr} and has since become the default in large-scale training, including the scaling-law studies that fixed much of current practice \citep{kaplan2020scaling}. Linear warmup was popularized for large-batch training \citep{goyal2017accurate}, cyclical policies that raise and lower the rate repeatedly were developed in parallel \citep{smith2017cyclical}, and the one-cycle policy and the super-convergence phenomenon grew out of that same line \citep{smith2019super}. A useful corrective comes from \citet{li2019exponential}, who showed that for scale-invariant networks an exponentially \emph{increasing} rate can be equivalent to a standard decaying schedule with weight decay, which is a reminder that the apparent shape of a schedule is not always what determines its effect. More recent work has been driven by a structural weakness of cosine annealing, namely that it must commit to the total training length in advance, so that extending a run requires recomputing the whole curve. The warmup-stable-decay schedule \citep{hu2024minicpm} holds the rate constant after warmup and decays only at the end, which decouples the decay from a predetermined step count, and \citet{wen2025understanding} give a loss-landscape account of why that late decay phase produces the sharp improvement it does. \citet{defazio2023optimal} argue on theoretical grounds for linear decay to zero, while \citet{defazio2024road} dispense with the schedule altogether in favor of a form of iterate averaging. This literature bears on our analysis in a way worth stating plainly: it seeks to reduce dependence on the horizon $E$, whereas the distinction we examine in Section~\ref{sec:brach} moves in the opposite direction and ties the schedule to $E$ more tightly. That the resulting effect vanishes as $E^{-2}$ is consistent with the view taken in that literature, namely that the endpoint of the decay matters chiefly through the closing phase of training. The present paper belongs to a smaller tradition of controlled empirical study rather than method proposal. \citet{gotmare2019closer} examined learning-rate restarts, warmup and distillation empirically and found that several of the standard justifications offered for these heuristics do not survive inspection. Our benchmark is narrower in scope but fully crossed, in that every schedule is run with every architecture on every dataset under one protocol, which is what allows the effect of the schedule to be separated from the effect of the setting. Against this background we make no claim to a new optimization mechanism. \method{} is cosine annealing with the half-period set to $E-1$, and Section~\ref{sec:brach} shows that the deviation this introduces from the conventional half-period $E$ shrinks quadratically in the horizon. What the paper contributes is an exact characterization of that deviation together with a controlled measurement, on a fixed protocol, of how much the choice of schedule is worth in the first place, relative to the choice made within the smooth family.
\section{Methodology}
\label{sec:method}

This section fixes the notation, defines \method{} and relates it analytically to cosine annealing, and
then sets out the baselines, architectures and training protocol against which it is evaluated. The
design principle throughout is that the schedule should be the only quantity that varies: the optimizer,
the initialization, the data pipeline and the evaluation criterion are held fixed across every run, so
that a difference in accuracy can be attributed to the shape of $\eta(e)$ alone. The comparison is fully
crossed rather than sampled, in that every schedule is run with every architecture on every dataset,
which keeps the design small enough to control closely and wide enough to show whether the behavior of a
schedule is consistent from one setting to the next.

\subsection{Setup and notation}
Every classifier is trained by minimizing the cross-entropy loss with the Adam optimizer
\citep{kingma2015adam}. Let $E$ denote the number of epochs, $e \in \{0,\dots,E-1\}$ the epoch index,
and $\etamax,\etamin$ the peak and floor learning rates. A schedule is a map from the epoch index to a
positive rate; the three smooth schedules considered here take values in $[\etamin,\etamax]$, whereas
the calendar-based rules follow their own multiplicative law and are not clipped to the floor. Every
schedule starts from $\etamax = 10^{-3}$, and $\etamin = 10^{-5}$ wherever a schedule has an explicit
floor, so that the differences we measure are differences of \emph{shape} rather than of range.

\subsection{The \method{} schedule}
\label{sec:brach}
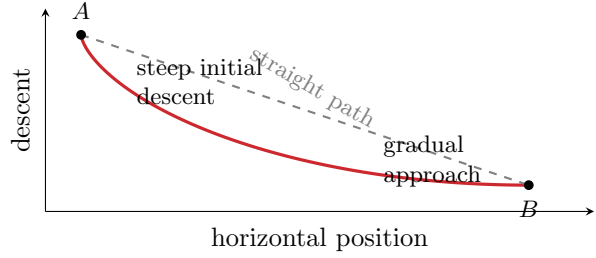
\begin{figure}[t]
\centering
\begin{tikzpicture}
\begin{axis}[
    width=0.58\textwidth, height=0.28\textwidth,
    xlabel={horizontal position}, ylabel={descent},
    xmin=-0.25, xmax=3.6, ymin=-2.35, ymax=0.35,
    xtick=\empty, ytick=\empty,
    axis lines=left, clip=false,
    label style={font=\small},
]
\addplot[domain=0:pi, samples=140, very thick, cBrach] ({x - sin(deg(x))}, {-(1 - cos(deg(x)))});
\addplot[dashed, gray, thick] coordinates {(0,0) (3.14159,-2)};
\node[circle,fill=black,inner sep=1.3pt,label={[font=\footnotesize]above:$A$}] at (axis cs:0,0) {};
\node[circle,fill=black,inner sep=1.3pt,label={[font=\footnotesize]below:$B$}] at (axis cs:3.14159,-2) {};
\node[font=\footnotesize,anchor=west,align=left] at (axis cs:0.32,-0.62) {steep initial\\ descent};
\node[font=\footnotesize,anchor=west,align=left] at (axis cs:2.05,-1.68) {gradual\\ approach};
\node[font=\footnotesize,gray,anchor=south,rotate=-30] at (axis cs:1.55,-0.95) {straight path};
\end{axis}
\end{tikzpicture}
\caption{The brachistochrone (curve of fastest descent, solid) between two points $A$ and $B$ is a
cycloid: it falls steeply at first and then flattens, in contrast to the straight path (dashed).
\method{} maps this descent profile onto the learning-rate schedule of \eqref{eq:brach}.}
\label{fig:cycloid}
\end{figure}

The brachistochrone is the path along which a body slides between two points in the least time under
gravity. Johann Bernoulli posed the problem in 1696, and the curve that solves it is a cycloid,
$x(\theta)=a(\theta-\sin\theta),\ y(\theta)=a(1-\cos\theta)$ (Figure~\ref{fig:cycloid}). Its descent is
steep at the start and gentle at the end, which is the profile one wants from a learning-rate schedule:
aggressive updates while the iterate is still far from a solution, small ones while it is being refined.
\method{} takes the normalized vertical coordinate of this cycloid and uses it to interpolate between
$\etamax$ and $\etamin$:
\begin{equation}
\eta(e) \;=\; \etamax - (\etamax - \etamin)\,\frac{1-\cos\theta}{2},
\qquad \theta \;=\; \frac{e}{E-1}\,\pi .
\label{eq:brach}
\end{equation}
At $e=0$ we have $\theta=0$ and $\eta=\etamax$; at $e=E-1$ we have $\theta=\pi$ and $\eta=\etamin$.
Figure~\ref{fig:lr} shows the schedule alongside the five baselines.

\paragraph{Relationship to cosine annealing.}
Expanding \eqref{eq:brach} gives the equivalent form
\begin{equation}
\eta(e) \;=\; \etamin + (\etamax-\etamin)\,\frac{1+\cos\!\big(\pi e/(E-1)\big)}{2},
\label{eq:brach-cos}
\end{equation}
which is \emph{exactly} cosine annealing with the half-period set to $E-1$ rather than $E$
\citep{loshchilov2017sgdr}. Two consequences follow, and we state both before drawing any conclusions
from the experiments. The first is practical: \eqref{eq:brach-cos} is what a standard cosine-annealing
implementation produces when its period argument is set to $E-1$ instead of $E$, so \method{} requires no
new code and is available in any current framework \citep{paszke2019pytorch}. The second is that the
brachistochrone should be read as motivation rather than as a derivation. The vertical coordinate of a
cycloid taken as a function of its parameter $\theta$ is a raised cosine by construction, and mapping
$\theta$ linearly onto the epoch index, as \eqref{eq:brach} does, is a choice: the curve itself is
$y$ as a function of $x$, and $x(\theta)=a(\theta-\sin\theta)$ is not linear in $\theta$, so a schedule
built from horizontal progress along the same curve would not be a cosine at all. The physical analogy is
therefore suggestive rather than load-bearing, and nothing in what follows depends on it.

The single structural consequence of the period choice is where the trajectory ends. \method{} is at
$\etamin$ at the last epoch that is actually trained, whereas cosine annealing
with half-period $E$ is still above the floor there, at
$\eta(E-1)=\etamin+(\etamax-\etamin)\big(1+\cos(\pi(E-1)/E)\big)/2$, and attains $\etamin$ only at the
index $e=E$, which is never reached. How much this matters depends on the horizon: since
$1+\cos\big(\pi(E-1)/E\big)=1-\cos(\pi/E)$, the excess above the floor is
$(\etamax-\etamin)\big(1-\cos(\pi/E)\big)/2\approx(\etamax-\etamin)\pi^{2}/(4E^{2})$, which for the
$E=10$ budget used here is $2.4\times10^{-5}$, larger than $\etamin$ itself, but for $E=100$ is
$2.4\times10^{-7}$, a perturbation of the floor of a few percent. \method{} thus decays slightly faster
than standard cosine annealing and finishes at a strictly lower rate, and the difference is one that can
matter for short training runs and becomes negligible for long ones; Sections~\ref{sec:results}
and~\ref{sec:discussion} examine what it is worth empirically. Beyond the horizon $E$ and the two
endpoints $\etamax$ and $\etamin$, the schedule has nothing to set: no decay milestones, no decay
factor, no warmup length, no restart period.

\subsection{Baseline schedules}
\begin{figure}[t]
\centering
\begin{tikzpicture}
\begin{semilogyaxis}[
    width=0.76\textwidth, height=0.35\textwidth,
    xlabel={epoch}, ylabel={learning rate},
    xmin=0, xmax=9, ymin=5e-7, ymax=2e-3,
    xtick={0,1,2,3,4,5,6,7,8,9},
    grid=both, grid style={gray!18},
    tick label style={font=\footnotesize},
    label style={font=\small},
    legend style={at={(1.02,1)}, anchor=north west, font=\footnotesize, draw=none},
    legend cell align=left,
]
\addplot[cConst, thick, dashed] coordinates
  {(0,1e-3)(1,1e-3)(2,1e-3)(3,1e-3)(4,1e-3)(5,1e-3)(6,1e-3)(7,1e-3)(8,1e-3)(9,1e-3)};
\addlegendentry{Constant}
\addplot[cStep, thick, const plot] coordinates
  {(0,1e-3)(1,1e-3)(2,1e-3)(3,1e-4)(4,1e-4)(5,1e-4)(6,1e-5)(7,1e-5)(8,1e-5)(9,1e-6)};
\addlegendentry{Step}
\addplot[cStep, only marks, mark=*, mark size=1.3pt, forget plot] coordinates {(9,1e-6)};
\addplot[cExp, thick] coordinates
  {(0,1e-3)(1,9.50e-4)(2,9.03e-4)(3,8.57e-4)(4,8.15e-4)(5,7.74e-4)(6,7.35e-4)(7,6.98e-4)(8,6.63e-4)(9,6.30e-4)};
\addlegendentry{Exponential}
\addplot[cCos, thick] coordinates
  {(0,1e-3)(1,9.66e-4)(2,9.06e-4)(3,7.96e-4)(4,6.58e-4)(5,5.05e-4)(6,3.52e-4)(7,2.14e-4)(8,1.00e-4)(9,3.43e-5)};
\addlegendentry{Cosine}
\addplot[cWarm, thick, dashdotted] coordinates
  {(0,3.0e-4)(1,1e-3)(2,9.63e-4)(3,8.55e-4)(4,6.94e-4)(5,5.05e-4)(6,3.16e-4)(7,1.55e-4)(8,4.77e-5)(9,1e-5)};
\addlegendentry{Warmup-cosine}
\addplot[cBrach, very thick] coordinates
  {(0,1e-3)(1,9.70e-4)(2,8.84e-4)(3,7.53e-4)(4,5.91e-4)(5,4.19e-4)(6,2.58e-4)(7,1.26e-4)(8,4.00e-5)(9,1e-5)};
\addlegendentry{\method{}}
\end{semilogyaxis}
\end{tikzpicture}
\caption{Learning-rate schedules over the $10$-epoch horizon (log scale, $\etamax=10^{-3}$,
$\etamin=10^{-5}$). Cosine annealing, warmup-cosine and \method{} share the smooth peak-to-floor
profile. Of the three, \method{} is the only one that both starts at $\etamax$ and is at $\etamin$ at
the final trained epoch: standard cosine annealing is still a factor of three above the floor when
training stops, while warmup-cosine reaches the floor only by compressing its cosine into the epochs
that follow the ramp.}
\label{fig:lr}
\end{figure}
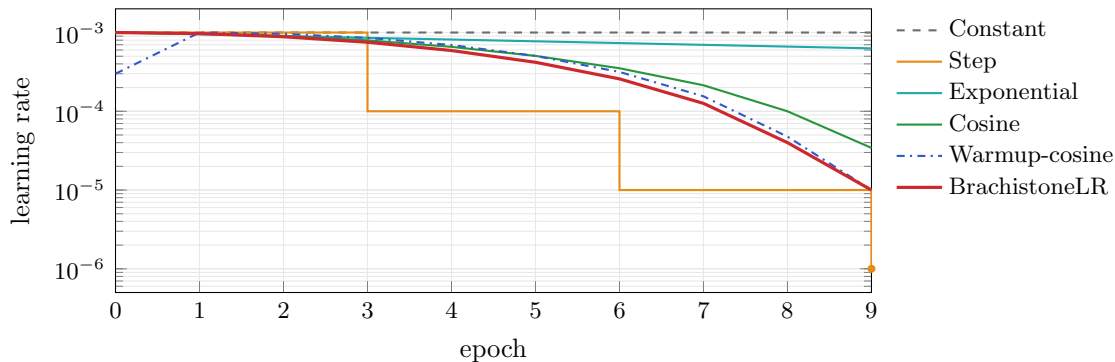

Table~\ref{tab:sched} gives the five baseline schedules with their update rules and the quantities each
one exposes. Together they cover the families in common use: no decay (constant), decay on a fixed
calendar (step, exponential), smooth peak-to-floor decay (cosine annealing), and a warmup followed by
smooth decay (warmup-cosine). Their trajectories over the $E=10$ horizon are drawn in
Figure~\ref{fig:lr}. Two of them do not respect the nominal range: with $\gamma=0.95$, exponential decay
has only fallen to $6.3\times10^{-4}$ by the last epoch and never approaches the floor, while step decay
with $\gamma=0.1$ and $s=3$ passes through the floor and ends an order of magnitude below it, at
$10^{-6}$. We report both as they are conventionally defined rather than clipping them, and return to
the point where it bears on the interpretation.

\begin{table}[t]
\centering
\small
\renewcommand{\arraystretch}{1.3}
\begin{tabular}{@{}lll@{}}
\toprule
\textbf{Schedule} & \textbf{Update rule $\eta(e)$} & \textbf{Hyperparameters} \\
\midrule
Constant           & $\etamax$ & - \\
Step decay         & $\etamax\,\gamma^{\lfloor e/s\rfloor}$ & $\gamma=0.1,\ s=\lfloor E/3\rfloor$ \\
Exponential decay  & $\etamax\,\gamma^{e}$ & $\gamma=0.95$ \\
Cosine annealing   & $\etamin+(\etamax-\etamin)\dfrac{1+\cos(\pi e/E)}{2}$ & - \\
Warmup-cosine      & linear ramp for $e<w$, then cosine & $w=\lceil 0.1E\rceil$ \\
\textbf{\method{} (ours)} & $\etamax-(\etamax-\etamin)\dfrac{1-\cos\theta}{2},\ \theta=\dfrac{\pi e}{E-1}$ & - \\
\bottomrule
\end{tabular}
\caption{The six learning-rate schedules compared. All start from $\etamax=10^{-3}$ and, where they have
an explicit floor, descend to $\etamin=10^{-5}$; only the shape varies. The two calendar-based rules are
not clipped to the floor, so exponential decay ends well above it and step decay one order of magnitude
below it.}
\label{tab:sched}
\end{table}

\subsection{Architectures}
We use four architecture families so that the comparison is not specific to a single inductive bias or
gradient-flow regime. The \emph{fully connected network} (FCN) stacks four dense layers
($512\text{-}256\text{-}128\text{-}10$) with batch normalization \citep{ioffe2015batch}, ReLU
activations, and dropout $p=0.3$ \citep{srivastava2014dropout}. The
\emph{convolutional network} (CNN) uses three convolutional blocks ($32\text{-}64\text{-}128$ filters,
$3\times3$ kernels) with batch normalization and $2\times2$ max-pooling, followed by a $256$-unit head.
The \emph{LSTM} \citep{hochreiter1997long} has two recurrent layers of $128$ units (dropout $0.3$) and
reads each image row by row as a sequence. The \emph{residual network} (ResNet) \citep{he2016deep} has
three residual blocks ($32\text{-}64\text{-}128$ channels) with skip connections, batch normalization,
and adaptive average pooling.

\subsection{Datasets and training protocol}
The three benchmarks are MNIST \citep{lecun1998gradient} ($70{,}000$ grayscale $28\times28$ digits),
Fashion-MNIST \citep{xiao2017fashion} ($70{,}000$ grayscale $28\times28$ garments) and CIFAR-10
\citep{krizhevsky2009learning} ($60{,}000$ RGB $32\times32$ objects), all ten-class problems. Images are
normalized with per-dataset channel statistics, and each training set is split $90/10$ into training and
validation by stratified sampling, with the official test set reserved for final evaluation. Every
configuration is trained with Adam ($\beta_1=0.9$, $\beta_2=0.999$, $\epsilon=10^{-8}$) from an initial
rate of $10^{-3}$ modulated by the schedule under test, using the cross-entropy loss, for $10$ epochs,
with batch sizes of $128$ for training and $256$ for evaluation. All $3\times4\times6=72$ configurations
start from identical initializations; we record accuracy and the learning rate at every epoch and report
the test accuracy of the epoch with the highest validation accuracy. The implementation is in PyTorch
\citep{paszke2019pytorch} and every run was performed on a single GPU. Each configuration was run once,
with a single random seed, so we treat sub-percentage-point differences as indicative and make no claims
of statistical significance (Section~\ref{sec:discussion}).

\section{Results}
\label{sec:results}

We report the twelve dataset--architecture combinations first at the level of individual configurations,
then in aggregate, and finally in terms of the patterns that depend on architecture and of the
trajectories themselves. The figure quoted for each configuration is the test accuracy at the epoch of
highest validation accuracy, as set out in Section~\ref{sec:method}, and every figure comes from a
single run. Small differences should therefore be read as indicative rather than as a strict ordering, a
caveat we take up in Section~\ref{sec:discussion}.

\subsection{Overall comparison}
Table~\ref{tab:main} gives the best test accuracy of each schedule on all twelve
dataset--architecture combinations. The three datasets discriminate to very different degrees: on MNIST
the six schedules occupy a band of $1.5$ points near the ceiling ($98.2$--$99.7\%$), on Fashion-MNIST
the band widens to $5.5$ points ($87.8$--$93.3\%$), and on CIFAR-10 it spans $30$ points
($53.4$--$83.4\%$). Counting best-in-row, cosine annealing leads on four combinations, \method{} and
warmup-cosine on three each, exponential decay on two, step decay on one, and the constant rate on none;
these counts sum to thirteen rather than twelve because cosine annealing and warmup-cosine tie on
Fashion-MNIST with the CNN. \method{} wins on MNIST with the LSTM and on both residual networks, and its
CIFAR-10 ResNet result ($83.38\%$) is both the best on that configuration, by $1.20$ points over cosine
annealing, and the highest single accuracy recorded anywhere on that dataset.

\begin{table}[t]
\centering
\small
\setlength{\tabcolsep}{5pt}
\begin{tabular}{@{}llcccccc@{}}
\toprule
\textbf{Dataset} & \textbf{Model} & Constant & Step & Exp. & Cosine & W-Cosine & \textbf{\method{}} \\
\midrule
\multirow{4}{*}{MNIST}
 & FCN    & 98.24 & 98.25 & 98.19 & \textbf{98.52} & 98.43 & 98.46 \\
 & CNN    & 99.23 & 99.50 & 99.35 & \textbf{99.52} & 99.46 & 99.46 \\
 & LSTM   & 98.92 & 98.80 & 98.94 & 99.03 & 99.02 & \textbf{99.07} \\
 & ResNet & 99.31 & 99.62 & 99.41 & 99.61 & 99.68 & \textbf{99.70} \\
\midrule
\multirow{4}{*}{Fashion-MNIST}
 & FCN    & 88.74 & 88.36 & 88.93 & 89.49 & \textbf{89.52} & 89.18 \\
 & CNN    & 91.96 & 92.61 & 92.34 & \textbf{92.84} & \textbf{92.84} & 92.54 \\
 & LSTM   & 89.01 & 87.79 & \textbf{89.46} & 89.41 & 89.39 & 89.25 \\
 & ResNet & 92.62 & \textbf{93.26} & 92.45 & 93.17 & 92.89 & 93.04 \\
\midrule
\multirow{4}{*}{CIFAR-10}
 & FCN    & 55.64 & 54.43 & 55.28 & \textbf{56.50} & 55.88 & 56.27 \\
 & CNN    & 74.84 & 75.69 & 77.36 & 77.51 & \textbf{78.26} & 77.68 \\
 & LSTM   & 54.92 & 53.36 & \textbf{55.82} & 55.41 & 55.66 & 55.50 \\
 & ResNet & 77.60 & 82.23 & 79.68 & 82.18 & 82.87 & \textbf{83.38} \\
\midrule
\multicolumn{2}{@{}l}{\textbf{\# best (of 12)}} & 0 & 1 & 2 & 4 & 3 & 3 \\
\bottomrule
\end{tabular}
\caption{Best test accuracy (\%) for each schedule across all $72$ configurations. The highest value in
each row is in bold; where two schedules tie, both are marked, which is why the counts in the last row
sum to thirteen rather than twelve. \method{} is best on three configurations, including both residual
networks, is among the three best schedules in ten of the twelve rows, and is never worse than fourth.}
\label{tab:main}
\end{table}

\subsection{Aggregate behavior}
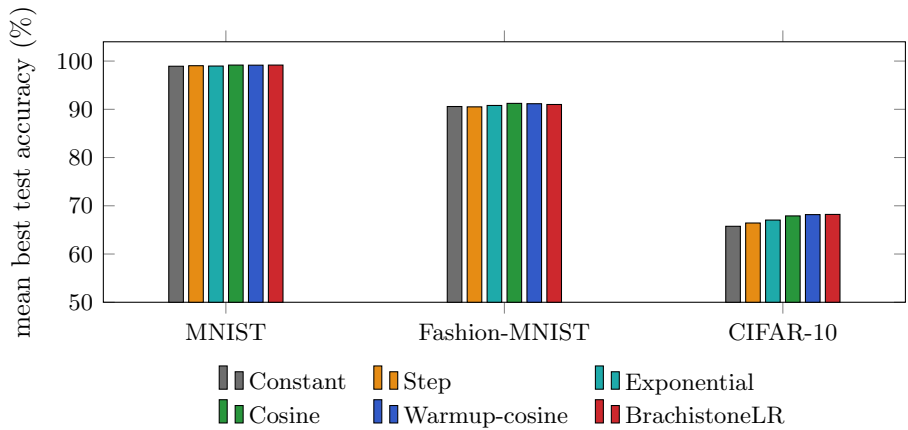
\begin{figure}[t]
\centering
\begin{tikzpicture}
\begin{axis}[
    ybar, width=0.80\textwidth, height=0.33\textwidth,
    bar width=5.5pt,
    enlarge x limits=0.22,
    ymin=50, ymax=104,
    ylabel={mean best test accuracy (\%)},
    symbolic x coords={MNIST, Fashion-MNIST, CIFAR-10},
    xtick=data,
    ytick={50,60,70,80,90,100},
    tick label style={font=\footnotesize},
    label style={font=\small},
    legend style={at={(0.5,-0.22)}, anchor=north, legend columns=3,
                  font=\footnotesize, draw=none, /tikz/every even column/.append style={column sep=8pt}},
    legend cell align=left,
]
\addplot[fill=cConst] coordinates {(MNIST,98.93)(Fashion-MNIST,90.58)(CIFAR-10,65.75)};
\addplot[fill=cStep]  coordinates {(MNIST,99.04)(Fashion-MNIST,90.51)(CIFAR-10,66.43)};
\addplot[fill=cExp]   coordinates {(MNIST,98.97)(Fashion-MNIST,90.80)(CIFAR-10,67.04)};
\addplot[fill=cCos]   coordinates {(MNIST,99.17)(Fashion-MNIST,91.23)(CIFAR-10,67.90)};
\addplot[fill=cWarm]  coordinates {(MNIST,99.15)(Fashion-MNIST,91.16)(CIFAR-10,68.17)};
\addplot[fill=cBrach] coordinates {(MNIST,99.17)(Fashion-MNIST,91.00)(CIFAR-10,68.21)};
\legend{Constant, Step, Exponential, Cosine, Warmup-cosine, \method{}}
\end{axis}
\end{tikzpicture}
\caption{Mean best test accuracy per dataset (averaged over the four architectures). Difficulty increases
from left to right. The three datasets share a common axis, so the differences \emph{between} schedules
within a dataset are compressed relative to the differences between datasets; the ordering is visible on
CIFAR-10, where the spread is largest, but Table~\ref{tab:agg} should be read for the values themselves.}
\label{fig:bars}
\end{figure}

Table~\ref{tab:agg} and Figure~\ref{fig:bars} aggregate over configurations. Three schedules stand apart
on both measures: cosine annealing ($86.10\%$ overall, mean rank $2.29$), warmup-cosine ($86.16\%$,
$2.42$) and \method{} ($86.13\%$, $2.46$). These three are separated by $0.06$ percentage points and
$0.17$ of a rank, while exponential ($85.60\%$), step ($85.33\%$) and constant ($85.09\%$) decay lie
half a point to a point behind and none of them achieves a mean rank better than $4.25$. The ordering is
stable across the three datasets, and on CIFAR-10, where the choice of schedule matters most, \method{}
has the highest dataset mean ($68.21\%$), just ahead of warmup-cosine ($68.17\%$) and cosine annealing
($67.90\%$).

\begin{table}[t]
\centering
\small
\setlength{\tabcolsep}{5.5pt}
\begin{tabular}{@{}lccccc@{}}
\toprule
\textbf{Schedule} & \textbf{MNIST} & \textbf{Fashion} & \textbf{CIFAR-10} & \textbf{Overall} & \textbf{Mean rank}$\downarrow$ \\
\midrule
Constant          & 98.93 & 90.58 & 65.75 & 85.09 & 5.33 \\
Step decay        & 99.04 & 90.51 & 66.43 & 85.33 & 4.25 \\
Exponential decay & 98.97 & 90.80 & 67.04 & 85.60 & 4.25 \\
Cosine annealing  & 99.17 & \textbf{91.23} & 67.90 & 86.10 & \textbf{2.29} \\
Warmup-cosine     & 99.15 & 91.16 & 68.17 & \textbf{86.16} & 2.42 \\
\textbf{\method{}}& \textbf{99.17} & 91.00 & \textbf{68.21} & 86.13 & 2.46 \\
\bottomrule
\end{tabular}
\caption{Mean best test accuracy (\%) per dataset and overall (averaged over the four architectures),
together with the mean rank across all twelve combinations (lower is better; ties receive the average
rank). \method{}, cosine annealing and warmup-cosine are separated by less than a tenth of a point
overall and jointly outperform the remaining three schedules. With one seed per configuration, the
ordering \emph{within} this leading group should not be read as meaningful. The best value in each
column is in bold.}
\label{tab:agg}
\end{table}

\subsection{Architecture and dataset effects}
The clearest architectural pattern concerns depth. \method{} is best on both residual networks (MNIST
$99.70\%$, CIFAR-10 $83.38\%$) and, averaged over the three datasets, has the highest ResNet accuracy of
any schedule ($92.04\%$, ahead of warmup-cosine at $91.81\%$ and step decay at $91.70\%$), which is
where a lower terminal rate has the most room to help. On convolutional networks it is never the best
but always close, finishing within $0.06$, $0.30$ and $0.58$ points of the row leader on MNIST,
Fashion-MNIST and CIFAR-10 respectively. The recurrent models are the exception: \method{} is best on
MNIST with the LSTM ($99.07\%$) but is edged out by exponential decay and warmup-cosine on the other two
datasets, and averaged over the three its LSTM accuracy is marginally below that of exponential decay
($81.27\%$ against $81.41\%$), which suggests that a higher sustained rate suits recurrent weight
updates. On fully connected networks the leading schedules sit within a fraction of a point of one
another (on MNIST, for instance, $98.46$ against $98.52$).

\subsection{Learning-rate trajectories}
Figure~\ref{fig:lr} makes the differences in shape concrete. The constant rate does not move;
exponential decay with $\gamma=0.95$ still retains $63\%$ of its initial value at the last epoch
($0.95^{9}\approx0.63$) and so never enters a low-rate regime; step decay falls in discrete factors of
ten; and cosine annealing, warmup-cosine and \method{} share the peak-to-floor descent. Of these three,
\method{} and warmup-cosine are the only ones that are at $\etamin$ when training stops, and \method{}
is the only one that arrives there without spending its first epoch below $\etamax$. Consistent with
Section~\ref{sec:brach}, this is the most plausible source of its small advantage on the deeper models
and the harder dataset. We also observed, in the per-epoch validation curves we recorded but do not
reproduce here, that the abrupt multiplicative drops of step decay were sometimes followed by a
transient loss of accuracy, whereas none of the smooth schedules produced such dips.

\section{Discussion}
\label{sec:discussion}

The benchmark separates the six policies along a single axis, the overall shape of the trajectory, and
is largely indifferent to how that shape is parameterized. The three schedules that fall smoothly from
the peak to the floor across the whole budget take the top three places in mean rank and finish within
$0.06$ points of one another, while the constant rate and the two calendar-based rules trail by half a
point to a point overall and never reach a mean rank better than $4.25$; the gap between the two groups
widens with the difficulty of the task, from a quarter of a point of dataset mean on MNIST to between
$0.9$ and $2.5$ points on CIFAR-10, and to $5.8$ points on the CIFAR-10 residual network, where
\method{} reaches $83.38\%$ and a constant rate reaches $77.60\%$. Two properties of the smooth policies
account for this. They hold the rate near $\etamax$ through the early and middle epochs, which is
consequential when the budget is only ten epochs long: exponential decay with $\gamma=0.95$ has
surrendered barely a third of its initial rate by the end of training and so never enters a genuine
fine-tuning regime, while step decay is at or below the floor from epoch six onwards and spends its last
four epochs at a rate from which little further progress is available. They also arrive at the small
rate continuously rather than in jumps, and the multiplicative drops of step decay were followed in our
runs by transient dips in validation accuracy that the smooth schedules never produced. Within the
leading group the argument has to be made more narrowly. \method{} differs from standard cosine
annealing in one respect only, the choice of half-period, and the consequence is that it sits at
$\etamin$ at the last epoch that is trained while cosine annealing is still at $3.4\times10^{-5}$, more
than three times the floor. The configurations on which \method{} wins outright are the ones on which
that difference would be expected to matter: the two residual networks, the deepest models in the study,
on which it also has the highest three-dataset mean of any schedule; and on the harder of the two the
three smooth schedules order themselves by terminal rate, with \method{} and warmup-cosine, both of
which finish at $\etamin$, taking the first two places and cosine annealing $1.20$ points behind. The
reading we favor is the conventional account of late-stage annealing, in which a final phase at a rate
that is genuinely small, rather than merely smaller than the peak, allows a deep model to settle within
a basin instead of continuing to traverse it. On that account a benefit should appear only where the
model is deep enough and the task hard enough that the closing epochs still carry information, which is
consistent with the absence of any effect on MNIST and on the shallower models. The same reasoning
bounds the claim in three ways. First, the advantage over standard cosine annealing is a function of the horizon and falls off as
$E^{-2}$: at $E=100$ the two schedules differ at the final epoch by a few percent of the floor and are
for practical purposes the same policy, so nothing here establishes \method{} as an improvement on
cosine annealing in general, only as the member of that family whose end point is aligned with the end
of training. Second, one seed per configuration cannot support an ordering within the leading group,
where $0.06$ points and $0.17$ of a rank separate the three schedules; \method{} is also fourth of six
on two configurations, both on Fashion-MNIST, and on the recurrent models its three-dataset mean is
marginally below that of exponential decay, so the advantage is neither uniform nor established, and
separating the top three with any confidence would require repeated runs with paired statistical tests.
Third, every schedule was run at a single shared peak rate of $10^{-3}$ with Adam. This is what makes the
comparison controlled, but it is also a confound: each policy has its own optimal peak, and a constant
rate in particular is never used untuned in practice, so the margin we report between the smooth
schedules and the constant baseline should be read as the margin \emph{at this peak} rather than the
margin that would survive tuning each policy separately. The same caveat covers the floor, fixed at
$10^{-5}$ throughout, the ten-epoch budget, and the coverage of the study: three small-to-medium vision
datasets and four moderate architectures, with SGD, longer horizons, larger models and other modalities
left open. It is also worth situating the result against the direction the literature has taken, which is
away from schedules that commit to a horizon at all: the warmup-stable-decay schedule holds the rate
constant and decays only at the end, decoupling the decay from a predetermined step count
\citep{hu2024minicpm,wen2025understanding}, and schedule-free methods dispense with the curve entirely
\citep{defazio2024road}. \method{} sits at the opposite extreme, binding the schedule to $E$ as tightly as
it can be bound, and our $E^{-2}$ result explains why that binding buys so little: the quantity it
controls, the gap between the terminal rate and the floor, is itself vanishing in the horizon. What the
results do support is a claim about cost rather than about ceilings. \method{} is better than three of
the five baselines on all three datasets and on three of the four architecture families, level with the
other two, and best on both residual networks and on the highest single CIFAR-10 result, and it obtains
this while exposing three quantities that are all fixed in advance by the training budget and the
optimizer. Since accuracy within the smooth family is for practical purposes the same, the sensible
criterion for choosing among its members is how much tuning each one demands, and by that criterion a
cosine whose half-period is the training horizon is the cheapest member and a reasonable default,
particularly for residual and convolutional models.

\section{Conclusion}
\label{sec:conclusion}

\method{} is a learning-rate schedule whose shape is borrowed from the curve of fastest descent and
which, once expanded, turns out to be cosine annealing with its half-period set to $E-1$, so that the
rate reaches its floor at the last epoch that is trained rather than one epoch later. We have been
explicit that this is a configuration of an existing schedule rather than a new one, and that the
deviation it introduces from the conventional half-period shrinks as $E^{-2}$, so it is a short-horizon
effect by construction. The benchmark that occupies most of the paper supports a more general conclusion
than the schedule itself does: across $72$ runs on a fixed protocol, the three smooth peak-to-floor
policies separate clearly from the constant rate and from calendar-based decay, by margins that grow with
task difficulty, while differing from one another by less than a tenth of an accuracy point. Within that
leading group \method{} is best on both residual networks and has the highest mean on CIFAR-10, but with
one seed per configuration we cannot and do not claim that this ordering is real. The practical reading is
that the choice between the smooth schedules is close to immaterial and the choice of whether to use one
at all is not. The obvious next steps follow from the limitations: repeated runs with paired significance
testing, a sweep over the horizon $E$ to test the predicted $E^{-2}$ decay directly, per-schedule tuning
of the peak rate, and comparison against schedules that do not fix the horizon in advance.

\vskip 0.2in
\bibliography{sample}

\end{document}